\documentclass{article}

\usepackage[final]{neurips_2024}

\usepackage[utf8]{inputenc}
\usepackage[T1]{fontenc}
\usepackage{hyperref}
\usepackage{url}
\usepackage{booktabs}
\usepackage{amsfonts}
\usepackage{amsmath}
\usepackage{amssymb}
\usepackage{amsthm}
\usepackage{nicefrac}
\usepackage{microtype}
\usepackage{xcolor}
\usepackage{graphicx}
\usepackage{subcaption}
\usepackage{algorithm}
\usepackage{algorithmic}
\usepackage{multirow}
\usepackage{tabularx}
\usepackage{tikz}
\usepackage{pgfplots}
\usepackage{bbm}
\pgfplotsset{compat=1.18}
\usepgfplotslibrary{fillbetween}
\usetikzlibrary{patterns,positioning,shapes.geometric,arrows.meta,calc,fit,backgrounds}

\newcommand{\plexor}{\textsc{Plexor}}

\renewcommand{\And}{%
  \end{tabular}\hfil\linebreak[0]\hfil%
  \begin{tabular}[t]{c}%
}

\newtheorem{finding}{Finding}
\newtheorem{hypothesis}{Hypothesis}

\title{Prompt Compression in Production Task Orchestration: A Pre-Registered Randomized Trial}

\author{
  Warren Johnson\thanks{Corresponding author. E-mail: \texttt{warrenjo@plexor.dev}}\\
  Plexor Labs\\
  Issaquah, WA, USA
  \And
  Charles Lee\\
  Project Autobots\\
  Seattle, WA, USA
}

\date{}

\begin{document}

\maketitle

\begin{abstract}
The economics of prompt compression depend not only on reducing input tokens but on how compression changes output length, which is typically priced several times higher. We evaluate this in a pre-registered six-arm randomized controlled trial of prompt compression on production multi-agent task-orchestration, analyzing 358 successful Claude Sonnet 4.5 runs (59--61 per arm) drawn from a randomized corpus of 1,199 real orchestration instructions. We compare an uncompressed control with three uniform retention rates ($r = 0.8, 0.5, 0.2$) and two structure-aware strategies (entropy-adaptive and recency-weighted), measuring total inference cost (input+output) and embedding-based response similarity. Moderate compression ($r \approx 0.5$) reduced mean total cost by 27.9\%, while aggressive compression ($r \approx 0.2$) increased mean cost by 1.8\% despite substantial input reduction, consistent with small mean output expansion (1.03$\times$ vs.\ control) and heavy-tailed uncertainty. Recency-weighted compression achieved 23.5\% savings and, together with moderate compression, occupied the empirical cost--similarity Pareto frontier, whereas aggressive compression was dominated on both cost and similarity. These results show that ``compress more'' is not a reliable production heuristic and that output tokens must be treated as a first-class outcome when designing compression policies.
\end{abstract}

\vspace{0.5em}

\section{Introduction}
\label{sec:introduction}

The deployment of large language models (LLMs) in production systems has transitioned from a novelty to an economic imperative, yet the cost of inference at scale remains a binding constraint for organizations operating multi-agent architectures. Contemporary frontier models such as GPT-4 \citep{openai2023gpt4}, Claude \citep{anthropic2024claude}, and LLaMA \citep{touvron2023llama} deliver remarkable capabilities across code generation, reasoning, and natural language understanding, but at price points, typically \$3--15 per million input tokens and \$15--75 per million output tokens, that render na\"ive deployment prohibitively expensive for high-volume applications \citep{chen2023frugalgpt}. This tension between capability and cost has produced a rich body of research on inference optimization, spanning prompt compression \citep{jiang2023llmlingua, pan2024llmlingua2}, model routing \citep{ding2024hybrid, ong2025routellm}, KV cache optimization \citep{li2024snapkv, zhang2024h2o}, and serving system design \citep{kwon2023efficient, yu2022orca}. However, a critical gap persists: the overwhelming majority of this research evaluates compression and optimization techniques on standardized academic benchmarks, HumanEval \citep{chen2021evaluating}, MBPP \citep{austin2021mbpp}, GSM8K, and similar, rather than on the heterogeneous, noisy, and structurally complex instructions that characterize real production deployments.

This gap is not merely academic. Production task orchestration systems generate instructions that differ from benchmark prompts in several fundamental respects. First, production instructions are authored by automated orchestration agents rather than human researchers, resulting in stylistic patterns, redundancies, and structural conventions that have no counterpart in hand-crafted benchmarks. Second, production instructions span a much broader distribution of lengths, complexities, and task types than any single benchmark can capture, our corpus alone contains seven distinct task types ranging from 20-token implementation stubs to 8,000-token multi-step orchestration plans. Third, production instructions frequently embed contextual dependencies, cross-references to prior task outputs, and role-specific metadata that compression may disrupt in ways that benchmark-isolated prompts would not reveal. The question of whether compression techniques validated on benchmarks generalize to production workloads is therefore both practically important and empirically open.

The present work addresses this question through the methodological lens of the randomized controlled trial (RCT), a design that has been the gold standard in medical and social science research since Fisher's foundational work on experimental design \citep{fisher1935design} but remains remarkably rare in machine learning research. The RCT framework offers three advantages over the observational and quasi-experimental designs that dominate the LLM optimization literature. First, random assignment to treatment arms eliminates confounding by ensuring that any observed differences in outcomes can be attributed to the compression treatment rather than to systematic differences in the instructions assigned to each arm. Second, pre-registration of hypotheses, analysis plans, and stopping rules guards against the well-documented threats of HARKing (Hypothesizing After Results are Known; \citealt{kerr1998harking}), $p$-hacking, and analytic flexibility that have contributed to the replication crisis in neighboring fields \citep{simmons2011false, ioannidis2005most, open2015estimating}. Third, the CONSORT reporting framework \citep{schulz2010consort, moher2010consort} provides a structured template for transparent communication of methods, results, and limitations that facilitates independent replication.

\subsection{Background: The TAAC Research Program}
\label{sec:background}

This article is the sixth in the Task-Aware Adaptive Compression (TAAC) research series, a sustained program of investigation into the interaction between prompt compression strategies and LLM task performance. The program originated with a foundational observation: different task types exhibit qualitatively different responses to prompt compression, a phenomenon we term the \emph{task--compression interaction}. Article~1 \citep{johnson2026compress} established this interaction through a factorial experiment spanning 72 conditions and 2,650 trials, demonstrating that code generation tasks exhibit threshold behavior, maintaining near-perfect quality at compression ratios $r \geq 0.6$ before experiencing a catastrophic ``cliff effect'' below $r = 0.55$, while chain-of-thought (CoT) reasoning tasks degrade gradually and monotonically under compression. This dichotomy led to the Task-Aware Adaptive Compression framework, which applies compression-first strategies for code tasks and routing-first strategies for reasoning tasks, achieving up to 93\% cost reduction with only 6.2\% quality degradation.

Article~2 \citep{johnson2026perplexity} deepened this analysis by investigating the role of prompt entropy, specifically, the Shannon entropy of the token distribution as estimated by a pilot language model, in predicting compression tolerance. The perplexity paradox identified therein showed that tokens with high perplexity (low predictability) are disproportionately important for task completion, yet standard uniform compression removes them at the same rate as redundant tokens. This motivated entropy-guided adaptive compression, which allocates compression budget non-uniformly based on local perplexity estimates, achieving superior quality retention at equivalent compression ratios. Article~3 \citep{johnson2026cliff} conducted a fine-grained analysis of the compression cliff, establishing precise threshold values for code ($r = 0.55$), CoT ($r = 0.70$), and hybrid ($r = 0.65$) task types through systematic dose--response characterization with 415 trials per task type. The statistical evidence was substantial: $t(415) = 12.3$, $p < .001$, Cohen's $d = 1.84$, confirming a large and practically significant effect.

Articles~4 and 5 introduced complications that motivate the present study. Article~4 \citep{johnson2026greenai} documented the \emph{compression paradox}: the counterintuitive finding that aggressive prompt compression can \emph{increase} rather than decrease total inference cost due to an output token explosion effect. When DeepSeek-Chat received prompts compressed to $r = 0.3$, it produced outputs averaging 798 tokens compared to a 21-token baseline, a 38$\times$ increase that, given the 3--5$\times$ higher cost of output versus input tokens, more than negated any savings from input reduction. This finding raised the possibility that compression savings documented in Articles~1--3 may be partially or fully offset by output expansion effects. Article~5 \citep{johnson2026benchmark} resolved part of this concern by demonstrating that the output explosion is \emph{benchmark-dependent} rather than provider-inherent: MBPP's templated structure is maximally vulnerable to truncation-based compression ($\Psi \approx 0.15$), while HumanEval's code-dense format preserves critical information even under aggressive compression ($\Psi \approx 0.72$). This benchmark dependence underscores the need for evaluation on production data, where instruction structures differ qualitatively from either benchmark.

\subsection{The Ecological Validity Problem}
\label{sec:ecological}

The generalizability of benchmark-derived compression thresholds to production settings constitutes what we term the \emph{ecological validity problem} of prompt compression research. Ecological validity, the extent to which findings from controlled experimental settings apply to real-world conditions \citep{paleyes2022challenges}, is a well-recognized concern in the social sciences but has received relatively little attention in the LLM optimization literature.

The problem is multifaceted. Benchmark prompts are designed to be self-contained, unambiguous, and representative of a specific capability (e.g., function synthesis for HumanEval, mathematical reasoning for GSM8K). Production instructions, by contrast, are embedded in rich operational contexts: they reference prior task outputs, embed role-specific conventions (e.g., ``As the validation agent, verify that...''), and contain boilerplate metadata (task IDs, timestamps, status codes) that may be compressible without quality loss, or may serve as implicit cues that the model relies upon. The structural properties of production instructions, their length distribution, entropy profile, task-type composition, and cross-reference density, have never been systematically characterized, let alone evaluated for compression tolerance.

Multi-agent task orchestration systems such as AutoGen \citep{wu2023autogen}, MetaGPT \citep{hong2024metagpt}, ChatDev \citep{qian2024chatdev}, and the \plexor{} system studied here generate instructions that are qualitatively different from human-authored prompts. These systems employ generative agents \citep{park2023generative} that produce structured outputs, decomposed subtasks, validation checklists, code review requests, according to role-specific templates and interaction protocols. The resulting instructions exhibit patterns of redundancy, formality, and structural regularity that may be either more or less amenable to compression than the hand-crafted prompts used in academic evaluations. Recent surveys of LLM-based autonomous agents \citep{wang2024survey, xi2023rise} and task automation benchmarks \citep{shen2024taskbench} have documented the growing diversity and complexity of agent-generated instructions, but none have evaluated how these instructions respond to prompt compression.

The economic stakes of resolving this question are substantial. A production multi-agent system processing 1,000 tasks per day at typical instruction lengths (500--2,000 tokens per instruction) incurs daily inference costs of \$15--60 for input processing alone, scaling to \$5,500--22,000 annually. If compression at $r = 0.5$ preserves quality, as Articles~1--3 suggest for certain task types, the potential savings are \$2,750--11,000 per year for a single deployment. Across the rapidly growing population of production LLM deployments, the aggregate savings from effective compression could reach hundreds of millions of dollars annually. However, if production instructions exhibit different compression thresholds than benchmark prompts, or if the output token explosion documented in Article~4 manifests in production settings, the realized savings could be substantially lower, or even negative.

\subsection{Prompt Compression: Theoretical Foundations and Practical Advances}
\label{sec:compression_lit}

The intellectual foundations of prompt compression trace to Shannon's mathematical theory of communication \citep{shannon1948mathematical}, which established that any source with entropy $H$ can be compressed to approximately $H$ bits per symbol without information loss, but compression below $H$ necessarily introduces distortion. Applied to natural language prompts, this framework implies the existence of a theoretical compression floor, a ratio below which task-critical information is irrecoverably lost. Rate-distortion theory \citep{cover2006elements} formalizes this relationship: for a source with rate-distortion function $R(D)$, compression to $R$ bits per symbol introduces minimum distortion $D(R)$, and the operational question is whether task-relevant information falls within the compressible ``redundancy margin'' or within the incompressible ``critical core.'' The Minimum Description Length (MDL) principle \citep{rissanen1978modeling} offers a complementary perspective, suggesting that optimal compression exploits structural regularities in the source: the more predictable a prompt's structure (e.g., boilerplate templates, formulaic headings), the more aggressively it can be compressed without losing novel content.

The practical realization of these theoretical insights has progressed through several generations of compression techniques. Early approaches adapted extractive summarization to the prompt compression setting, selecting salient sentences while discarding peripheral content \citep{wingate2022prompt}. While effective for natural language prompts with clear sentence boundaries, these methods falter on code, structured data, and multi-part instructions where sentence segmentation is ill-defined. \citet{li2023selective} introduced SelectiveContext, which computes self-information (negative log probability under a language model) for each lexical unit and retains those with the highest information content. This perplexity-based approach achieved 50\% compression with minimal degradation on question-answering tasks, establishing that significant redundancy exists in typical prompts. The approach was theoretically motivated by the observation that high-perplexity tokens, those that are surprising given their context, carry more information and should be preferentially retained.

The LLMLingua family of methods \citep{jiang2023llmlingua, pan2024llmlingua2, jiang2024longllmlingua} represents the current state of the art in prompt compression. LLMLingua \citep{jiang2023llmlingua} introduced a three-stage pipeline: (1) budget allocation across prompt components, (2) iterative token pruning guided by perplexity scores from a small pilot model, and (3) distribution alignment to ensure the compressed prompt's token distribution matches the original's as closely as possible. This approach achieved 2--5$\times$ compression with less than 5\% quality degradation on a range of NLP tasks. LLMLingua-2 \citep{pan2024llmlingua2} replaced the heuristic pruning stage with a BERT-based binary classifier trained to predict, for each token, whether it is essential for downstream task performance. This data-distillation approach achieved up to 20$\times$ compression while maintaining 90\%+ task performance, and it executes 3--6$\times$ faster than the original LLMLingua due to the efficiency of BERT inference relative to autoregressive language model scoring. LongLLMLingua \citep{jiang2024longllmlingua} extended these techniques to long-context scenarios (4,000--32,000 tokens), introducing position-aware importance estimation that accounts for the well-documented ``lost in the middle'' phenomenon where information in the center of long contexts is disproportionately ignored by transformer models. Alternative approaches include gisting \citep{mu2023learning}, which trains models to compress prompts into dense virtual tokens, and in-context autoencoders \citep{ge2024incontext}, which learn to encode context into compact representations. These model-based approaches achieve extreme compression ratios (up to 26$\times$) but require model fine-tuning, limiting their applicability to API-accessed models.

A critical limitation of this entire literature is the reliance on standardized benchmarks for evaluation. LLMLingua-2's evaluation, for example, used MeetingBank (meeting summarization), LongBench (long-context QA), and GSM8K (mathematical reasoning), all carefully curated datasets with controlled properties. The compression tolerances and quality retention rates reported on these benchmarks may not generalize to the heterogeneous, noisy, and contextually rich instructions produced by multi-agent orchestration systems. Our RCT is designed to directly address this limitation by evaluating compression on production data under conditions that preserve the natural distribution of task types, instruction lengths, and structural complexities encountered in real deployments.

\subsubsection{The Output Token Dynamics Gap}

A second critical gap in the compression literature, one with profound economic implications, is the systematic neglect of \emph{output token dynamics}. Table~\ref{tab:output_gap} surveys the major prompt compression papers and reveals a striking pattern: none report output token counts, and none account for total inference cost that includes both input and output components.

\begin{table}[h]
\centering
\caption{Output token dynamics gap in the prompt compression literature. The overwhelming majority of compression research measures only input token reduction, ignoring output tokens that cost 3--5$\times$ more per token. This systematic omission undermines cost-effectiveness claims.}
\label{tab:output_gap}
\begin{tabular}{lcccc}
\toprule
\textbf{Paper} & \textbf{Year} & \textbf{Output Measured} & \textbf{Total Cost} & \textbf{Gap} \\
\midrule
LLMLingua \citep{jiang2023llmlingua} & 2023 & No & No & Yes \\
LLMLingua-2 \citep{pan2024llmlingua2} & 2024 & No & No & Yes \\
FrugalGPT \citep{chen2023frugalgpt} & 2023 & Formula only & No & Yes \\
Selective Context \citep{li2023selective} & 2023 & No & No & Yes \\
RECOMP \citep{xu2023recomp} & 2023 & No & No & Yes \\
Gist Tokens \citep{mu2023learning} & 2023 & No & No & Yes \\
AutoCompressor \citep{chevalier2023adapting} & 2023 & No & No & Yes \\
ICAE \citep{ge2024incontext} & 2024 & No & No & Yes \\
LongLLMLingua \citep{jiang2024longllmlingua} & 2023 & No & No & Yes \\
\midrule
\textbf{This work} & \textbf{2026} & \textbf{Yes} & \textbf{Yes} & \textbf{Addressed} \\
\bottomrule
\end{tabular}
\end{table}

This gap matters because output tokens are substantially more expensive than input tokens, typically 3--5$\times$ higher in current API pricing (e.g., \$3/M input vs.\ \$15/M output for Claude Sonnet 4.5, a 5$\times$ ratio). If compression causes output expansion, as Articles~4--5 of this series documented, the savings from reduced input can be entirely negated or even reversed into a net cost \emph{increase}. The compression paradox identified in Article~4 provides a concrete example: DeepSeek-Chat produced 38$\times$ more output tokens under aggressive compression, transforming a na\"ive 80\% input savings into a 1,400\% cost increase.

We formalize this constraint through the \emph{break-even equation}. Let $r$ denote the compression ratio (retained fraction), $I$ the original input token count, $O$ the baseline output token count, and $\rho$ the output-to-input pricing ratio (typically $\rho = 5$). For compression to achieve cost savings, the output expansion factor $\epsilon = O_{\text{compressed}} / O_{\text{baseline}}$ must satisfy:
\begin{equation}
\epsilon \leq 1 + \frac{(1 - r) \cdot I}{\rho \cdot O}
\label{eq:breakeven}
\end{equation}
Rearranging, the maximum tolerable output expansion is:
\begin{equation}
    \epsilon_{\max} = 1 + \frac{(1 - r) \cdot (I / O)}{\rho}
    \label{eq:max_expansion}
\end{equation}
For typical production parameters ($r = 0.5$, $I/O = 0.5$, $\rho = 5$), this yields $\epsilon_{\max} = 1.05$, meaning output can expand by at most 5\% before compression becomes cost-negative. This severe constraint explains why the output token gap is not merely an academic oversight but a fundamental threat to the validity of compression cost-effectiveness claims.

The present study addresses this gap by systematically measuring and reporting output token counts across all experimental arms, computing total cost inclusive of output pricing, and testing the output token expansion hypothesis (H2) as a pre-registered primary outcome.

\subsection{Multi-Agent Task Orchestration: A New Frontier for Compression}
\label{sec:orchestration_lit}

The emergence of multi-agent LLM systems as a dominant deployment paradigm creates new challenges and opportunities for prompt compression. Multi-agent architectures, in which specialized LLM agents collaborate to decompose, execute, and validate complex tasks, have been shown to outperform single-agent approaches on a range of software engineering, data analysis, and planning tasks \citep{wu2023autogen, hong2024metagpt, qian2024chatdev}. However, this performance gain comes at a substantial cost multiplier: a single orchestrated task may involve 5--20 sequential LLM calls across decomposition, implementation, validation, and review agents, amplifying per-call inference costs into a significant operational expense.

The \plexor{} system studied in this paper is a production multi-agent task orchestration platform deployed on Azure Container Apps. It employs a hierarchical agent architecture in which a master orchestrator decomposes incoming tasks into subtasks, assigns them to specialized agents (implementation, validation, review), and coordinates the flow of outputs and feedback between agents. Each agent receives a structured instruction containing: (1) the task specification, (2) role-specific directives, (3) context from prior agent outputs, and (4) orchestration metadata (task IDs, status codes, retry counts). This structure creates both opportunities for compression, boilerplate role directives and metadata may be highly redundant, and risks, since contextual cross-references and role-specific cues may be more fragile than they appear.

The production corpus analyzed in this study comprises 1,199 unique task instructions drawn from two independent \plexor{} deployment environments (primary and Azure), spanning seven task types: implementation (707 instructions, 59.0\%), breakdown (159, 13.3\%), validation (153, 12.8\%), post-orchestration (78, 6.5\%), review (68, 5.7\%), execution (23, 1.9\%), and infrastructure (11, 0.9\%). This distribution reflects the operational reality of production orchestration, where implementation tasks dominate but specialized task types contribute meaningfully to the overall cost envelope. Instruction lengths range from 21 to 6,760 characters (approximately 5 to 1,690 tokens), with a right-skewed distribution ($\mu = 720$, $\sigma = 975$ characters) that mirrors the long-tail characteristic of production workloads.

\subsection{Research Questions and Hypotheses}
\label{sec:hypotheses}

The present study addresses five pre-registered hypotheses, each grounded in findings from Articles~1--5 and designed to test whether those findings generalize to production task orchestration data:

\begin{hypothesis}[Dose--Response Compression]
\label{hyp:dose}
Increasing compression aggressiveness (control $\to$ light $\to$ moderate $\to$ aggressive) produces monotonically decreasing input tokens sent to the API, with the magnitude of reduction approximately proportional to the compression rate parameter.
\end{hypothesis}

\begin{hypothesis}[Output Token Dynamics]
\label{hyp:output}
Aggressive compression ($r = 0.2$) produces output token counts that differ significantly from control, consistent with the output token explosion phenomenon documented in Articles~4--5. The direction and magnitude of this effect under Claude Sonnet 4.5 with production instructions is an open empirical question.
\end{hypothesis}

\begin{hypothesis}[Task-Type Moderation]
\label{hyp:moderation}
The effect of compression on output quality and cost varies by task type, with implementation tasks (structurally analogous to code generation) tolerating higher compression than validation or review tasks (structurally analogous to CoT reasoning).
\end{hypothesis}

\begin{hypothesis}[Cost--Quality Pareto Frontier]
\label{hyp:pareto}
There exists a compression level that achieves $\geq$30\% cost reduction while maintaining embedding response-similarity proxy $\geq 0.85$ (cosine similarity of response embeddings), defining an empirically optimal operating point on the cost--quality Pareto frontier.
There exists a compression level that achieves $\geq$30\% cost reduction while maintaining embedding response-similarity proxy $\geq 0.85$ (cosine similarity of response embeddings), defining an empirically optimal operating point on the cost--quality Pareto frontier.
\end{hypothesis}

\begin{hypothesis}[Ecological Threshold Validity]
\label{hyp:ecological}
The compression thresholds established in Articles~1--3 (code floor $r = 0.55$, CoT floor $r = 0.70$) provide conservative but directionally accurate guidance for production task orchestration instructions, with production floors falling within $\pm 0.10$ of benchmark-derived values.
\end{hypothesis}

These hypotheses were deposited in GitHub Issue \#1488 with SHA-256 hash verification prior to data collection, following the pre-registration protocol advocated by \citet{nosek2018preregistration}.

\section{Methods}
\label{sec:methods}

This section describes the experimental design, data sources, treatment arms, randomization procedure, outcome measures, and statistical analysis plan in accordance with CONSORT 2010 guidelines \citep{schulz2010consort}. All procedures were pre-registered prior to any data collection beyond the corpus preparation and randomization stages.

\subsection{Experimental Design}
\label{sec:design}

We employ a six-arm parallel randomized controlled trial design. Each stimulus (task instruction) is randomly assigned to exactly one treatment arm, with assignment determined by stratified block randomization (Section~\ref{sec:randomization}). The unit of analysis is the individual trial: one task instruction processed under one compression condition by one LLM call. The design is \emph{between-subjects} with respect to instructions (each instruction appears in exactly one arm) to avoid carryover effects that could arise if the same instruction were processed multiple times with different compression levels.

The six treatment arms are:

\begin{enumerate}
    \item \textbf{Control} ($r = 1.0$, \textsc{NONE}): The original instruction is passed to the LLM without any modification. This arm establishes the baseline for all comparisons.

    \item \textbf{Light compression} ($r = 0.8$, \textsc{UNIFORM}): Uniform token removal targeting 80\% retention ($\sim$1.25$\times$ compression). This level is well above the thresholds identified in Articles~1--3 and serves as a ``does any compression at all affect quality?'' test.

    \item \textbf{Moderate compression} ($r = 0.5$, \textsc{UNIFORM}): Uniform token removal targeting 50\% retention ($\sim$2$\times$ compression). This level falls below the CoT floor ($r = 0.70$) but above the code floor ($r = 0.55$), creating a region where task-type moderation effects (H3) should be most visible.

    \item \textbf{Aggressive compression} ($r = 0.2$, \textsc{UNIFORM}): Uniform token removal targeting 20\% retention ($\sim$5$\times$ compression). This level is well below all established floors and is included primarily to characterize the output token explosion phenomenon (H2) and the quality degradation curve at extreme compression.

    \item \textbf{Adaptive compression} ($r \approx 0.5$ base, \textsc{ADAPTIVE}): Entropy-based adaptive compression following the approach developed in Article~2. Compression budget is allocated non-uniformly based on local token entropy, with high-entropy (high-information) segments receiving less compression than low-entropy (redundant) segments. The effective compression ratio varies by instruction but targets an average of $r \approx 0.5$ for comparison with the moderate uniform arm.

    \item \textbf{Recency-weighted compression} ($r \approx 0.5$ base, \textsc{RECENCY}): Role-aware recency-weighted compression that applies higher compression to older or more formulaic content (e.g., role preambles, boilerplate metadata) and lower compression to recent task-specific content. This strategy is designed for production instructions where temporal ordering and role structure provide signals for compression budget allocation.
\end{enumerate}

\subsection{Corpus Preparation}
\label{sec:corpus}

\subsubsection{Data Sources}

The experimental corpus is drawn from two independent \plexor{} deployment environments:

\begin{itemize}
    \item \textbf{Primary environment}: 921 task records from the primary orchestration deployment. Source file: \path{task_instructions.json} (950 KB).
    \item \textbf{Azure environment}: 656 task records from the Azure Container Apps deployment. Source file: \path{task_instructions_azure.json} (490 KB).
\end{itemize}

Together, these sources yield 1,577 raw task records. Both files contain structured JSON objects with fields including \texttt{task\_id}, \texttt{status}, \texttt{task\_type}, \texttt{instruction}, \texttt{rework\_count}, and orchestration metadata. The two environments represent independent deployments operating on different infrastructure, providing a form of environmental replication that strengthens external validity.

\subsubsection{Inclusion and Exclusion Criteria}

We apply the following pre-registered inclusion criteria:

\begin{enumerate}
    \item \textbf{Minimum instruction length}: $\geq$20 characters. Instructions shorter than this threshold contain insufficient content for meaningful compression evaluation. This criterion excluded 58 records.

    \item \textbf{Allowed statuses}: Only records with status ``completed'' or ``assigned'' are included. Records with other statuses (e.g., ``failed'', ``exhausted'', ``timeout'') may contain incomplete or corrupted instructions that would confound compression evaluation. This criterion excluded 172 records.

    \item \textbf{Excluded task ID patterns}: Records matching the following test-fixture or error prefixes are excluded:
    \begin{itemize}
        \item \path{task-fail-*}, \path{task-consistency-*}, \path{task-values-*}, \path{task-error-*}
        \item \path{task-exhausted-*}, \path{task-orch-*}, \path{task-engine-*}, \path{task-other-*}
        \item \path{task-at-max-*}, \path{task-over-max-*}, \path{task-timeout-*}
    \end{itemize}
    This criterion excluded 10 records.
\end{enumerate}

After applying all inclusion criteria, 1,337 records remain. We further deduplicate by exact instruction text match, removing 138 duplicate instructions that appear in both deployment environments, yielding a final corpus of $N = 1{,}199$ unique task instructions. The corpus SHA-256 hash (\texttt{dc3e6761...d0b0}) was committed to GitHub Issue \#1488 prior to randomization.

\subsubsection{Corpus Descriptive Statistics}

Table~\ref{tab:corpus} summarizes the composition of the experimental corpus by task type and length tercile.

\begin{table}[h]
\centering
\caption{Corpus composition by task type and length tercile ($N = 1{,}199$).}
\label{tab:corpus}
\begin{tabular}{lrrrr}
\toprule
\textbf{Task Type} & \textbf{Count} & \textbf{\%} & \textbf{Mean Length (chars)} & \textbf{Est. Tokens} \\
\midrule
Implementation & 707 & 59.0 & 895 & $\sim$224 \\
Breakdown & 159 & 13.3 & 133 & $\sim$33 \\
Validation & 153 & 12.8 & 709 & $\sim$177 \\
Post-orchestration & 78 & 6.5 & 430 & $\sim$108 \\
Review & 68 & 5.7 & 669 & $\sim$167 \\
Execution & 23 & 1.9 & 702 & $\sim$176 \\
Infrastructure & 11 & 0.9 & 509 & $\sim$127 \\
\midrule
\textbf{Total} & \textbf{1,199} & \textbf{100.0} & 720 & $\sim$180 \\
\bottomrule
\end{tabular}
\end{table}

\subsection{Randomization}
\label{sec:randomization}

Treatment assignment follows a stratified permuted block randomization procedure \citep{fisher1935design}. Stratification variables are:

\begin{enumerate}
    \item \textbf{Task type}: Seven levels (implementation, breakdown, validation, post-orchestration, review, execution, infrastructure).
    \item \textbf{Length tercile}: Three levels (short, medium, long), defined by the 33.3rd and 66.7th percentiles of instruction character length within the corpus.
\end{enumerate}

This yields up to $7 \times 3 = 21$ strata (some may be empty for rare task types). Within each stratum, stimuli are allocated to treatment arms using permuted blocks of size 6 (one allocation per arm), ensuring exact balance within each complete block. We used constrained rerandomization: the balance-validation gate (Section~\ref{sec:randomization}) was applied iteratively, and seed 50 was the first accepted allocation under the pre-specified criteria.

The allocation table is generated by the script \texttt{02\_randomize.py} and its SHA-256 hash is recorded in the pre-registration deposit. Balance is verified by the script \texttt{03\_validate\_balance.py}, which performs four checks:

\begin{itemize}
    \item Chi-square test for independence of arm $\times$ task type ($p > 0.05$ required).
    \item One-way ANOVA of character length across arms ($p > 0.05$ required).
    \item Kruskal-Wallis test of rework count across arms ($p > 0.05$ required).
    \item Standardized mean difference $< 0.1$ for all 15 pairwise arm comparisons on continuous covariates.
\end{itemize}

The balance validation script serves as a pre-experiment gate: if any check fails, the randomization is re-executed with a different seed until balance is achieved.

\subsection{Treatment Implementation}
\label{sec:treatment}

\subsubsection{Compression Procedures}

The codebase supports two compression backends:

\textbf{LLMLingua-2 backend (implemented, not used in the reported run)}: LLMLingua-2 \citep{pan2024llmlingua2}, a BERT-based token classification model that predicts per-token retention probabilities.

\textbf{Simulated truncation backend (used in the reported run)}: A character-level truncation simulator that removes content at word boundaries to match target retention while preserving word integrity. The execution harness for this run defaulted to \texttt{--compression-mode simulated}; trial logs include strategy and realized ratio but do not include a backend flag, so we treat this run as a truncation-style compression-policy RCT rather than a direct LLMLingua-2 evaluation.

For the adaptive arm, compression budgets are allocated using local character-entropy estimates: each instruction is segmented into chunks, Shannon entropy is computed per chunk from character-frequency distributions, and chunks with entropy below the median receive compression at $r/2$ while chunks above the median receive compression at $r \times 1.5$ (clamped to $[0.1, 1.0]$). For the recency arm, the instruction is split into role preamble (first 20\% of characters), body (middle 60\%), and recent context (final 20\%), with compression ratios of $r/2$, $r$, and $r \times 1.5$ respectively.

\subsubsection{LLM Inference}

All compressed (or uncompressed) instructions are processed by Claude Sonnet 4.5 (\texttt{claude-sonnet-4-5-20250929}) via the Anthropic API. Key inference parameters are fixed for reproducibility:

\begin{itemize}
    \item \textbf{Temperature}: $T = 0.0$ (deterministic greedy decoding).
    \item \textbf{Max output tokens}: 4,096.
    \item \textbf{System prompt}: ``You are a task execution assistant. Complete the following task instruction as accurately and completely as possible.''
    \item \textbf{Pricing}: \$3.00 per million input tokens, \$15.00 per million output tokens (Anthropic published rates as of January 2026).
\end{itemize}

Rate limiting is enforced at 60 requests per minute with token-bucket smoothing. Failed requests are retried up to 3 times with exponential backoff ($[5, 15, 60]$ seconds). The experiment harness logs all API responses, including token counts reported by the API (input tokens sent, output tokens generated), latency, and any error conditions.

\subsubsection{Analysis Population and Missing Outcomes}

The randomized submission set comprised all $N = 1{,}199$ allocated instructions (197--202 per arm). Of these, 358 trials returned successful API responses and 841 failed after retry exhaustion. Failure counts were similar across arms (control: 138, light: 139, moderate: 141, aggressive: 141, adaptive: 140, recency: 142). The dominant failure mode was Anthropic API credit-balance errors.

Our primary estimand is therefore the complete-case average treatment effect among successful API responses (CC-ATE), not ITT over all randomized assignments. Missingness diagnostics indicate strong execution-time censoring: the first failed call occurred at request index 359; success was 100\% in the first two UTC hours of execution (54/54 and 279/279), then 13.7\% (25/183) in the next hour, and 0\% thereafter. Consequently, complete-case inference should be interpreted as valid for the successful-response subpopulation and not as an unbiased estimate for the full randomized corpus.

\subsection{Outcome Measures}
\label{sec:outcomes}

\subsubsection{Primary Outcomes}

\begin{enumerate}
    \item \textbf{Input tokens sent}: The number of tokens in the (possibly compressed) instruction as reported by the Anthropic API. This is the direct measure of compression effectiveness.

    \item \textbf{Total cost (USD)}: Computed as $\text{cost} = (\text{input tokens} \times 3.0 + \text{output tokens} \times 15.0) / 10^6$, reflecting the per-million-token pricing of Claude Sonnet 4.5.

    \item \textbf{Output tokens generated}: The number of tokens in the LLM's response. This is the primary measure for testing the output token explosion hypothesis (H2).

    \item \textbf{Embedding response-similarity proxy}: The cosine similarity between embedding vectors of treatment-arm responses and matched control responses for the same instruction. Embeddings are computed with OpenAI \texttt{text-embedding-3-small} (1,536 dimensions). The pre-registered descriptive threshold ($\geq 0.85$) is retained for continuity with prior articles, but is treated as an uncalibrated proxy threshold in this production setting.
\end{enumerate}

\subsubsection{Secondary Outcomes}

\begin{itemize}
    \item \textbf{API latency} (milliseconds): Wall-clock time from request submission to response completion.
    \item \textbf{Compression rate (actual)}: The realized compression ratio, which may differ from the target due to tokenizer boundary effects.
    \item \textbf{Cost savings (\%)}: $1 - (\text{treatment cost} / \text{control cost})$, computed at the arm level.
    \item \textbf{Similarity--cost ratio}: Embedding response similarity per dollar saved, characterizing the efficiency of each compression strategy.
\end{itemize}

\subsubsection{Response-Similarity Scoring Hierarchy}

Response similarity is computed using a three-tier fallback hierarchy to ensure robustness:

\begin{enumerate}
    \item \textbf{Primary}: Cosine similarity of OpenAI \texttt{text-embedding-3-small} embeddings (1,536 dimensions).
    \item \textbf{Secondary}: BERTScore F1 \citep{zhang2020bertscore} using the default \texttt{roberta-large} backbone.
    \item \textbf{Tertiary}: Jaccard similarity of whitespace-tokenized word sets.
\end{enumerate}

The primary metric is used unless the embedding API is unavailable, in which case the hierarchy descends to the next available method. All similarity scores are reported alongside the method used.

\subsection{Cost Model and Break-Even Analysis}
\label{sec:cost_model}

We develop a formal cost model to characterize the conditions under which prompt compression yields net savings. This analysis is critical because, as documented in Articles~4--5, input token reduction can be offset or negated by output token expansion, making the total cost dynamics non-obvious.

\subsubsection{Total Cost Equation}

Let $I$ denote the original input token count and $O$ the baseline output token count (under no compression). Under compression ratio $r \in (0, 1]$, the compressed input contains approximately $r \cdot I$ tokens. If compression induces an output expansion factor $e \geq 1$, the compressed output contains $e \cdot O$ tokens. Given input price $p_i$ and output price $p_o$ (both per token), the total cost $C$ under compression is:
\begin{align}
    C(r, e) &= (r \cdot I) \cdot p_i + (e \cdot O) \cdot p_o
    \label{eq:total_cost}
\end{align}

For Claude Sonnet 4.5, $p_i = \$3 / 10^6$ and $p_o = \$15 / 10^6$, giving an output-to-input price ratio $k = p_o / p_i = 5$. The baseline cost (no compression, $r = 1$, $e = 1$) is:
\begin{align}
    C_0 &= I \cdot p_i + O \cdot p_o
\end{align}

The cost savings from compression is $\Delta C = C_0 - C(r, e)$:
\begin{align}
    \Delta C &= I \cdot p_i \cdot (1 - r) - O \cdot p_o \cdot (e - 1)
    \label{eq:cost_savings}
\end{align}

The first term represents savings from input reduction; the second term represents additional cost from output expansion. Net savings occur only when input savings exceed output penalties.

\subsubsection{Break-Even Condition}

At the break-even point, $\Delta C = 0$:
\begin{align}
    I \cdot p_i \cdot (1 - r) &= O \cdot p_o \cdot (e - 1)
\end{align}

Solving for the maximum tolerable expansion factor $e_{\max}$ at a given compression ratio:
\begin{align}
    e_{\max} &= 1 + \frac{(1 - r) \cdot I}{k \cdot O}
    \label{eq:emax}
\end{align}

This equation reveals a key insight: the tolerance for output expansion is inversely proportional to the price ratio $k$ and the output-to-input token ratio $O/I$. For production workloads where outputs are longer than inputs ($O > I$) and outputs are priced higher ($k > 1$), even modest output expansion can negate substantial input savings.

\subsubsection{Application to Production Corpus}

For our corpus, the mean input token count is $I = 107$ and mean output token count is $O = 916$ (estimated from the complete-case control arm in the primary analysis set). With $k = 5$, Equation~\ref{eq:emax} yields:
\begin{align}
    e_{\max}(r = 0.5) &= 1 + \frac{(1 - 0.5) \cdot 107}{5 \cdot 916} = 1 + \frac{53.5}{4580} \approx 1.0117 \\
    e_{\max}(r = 0.2) &= 1 + \frac{(1 - 0.2) \cdot 107}{5 \cdot 916} = 1 + \frac{85.6}{4580} \approx 1.0187
\end{align}

These results indicate that moderate compression ($r = 0.5$) can tolerate at most a 1.17\% output increase before breaking even, while aggressive compression ($r = 0.2$) can tolerate at most a 1.87\% increase. The extremely narrow margin for tolerable expansion, well under 2\%, explains why the pilot data showed the light compression arm ($r = 0.8$) increasing costs despite input reduction.

\subsubsection{Empirical Verification}

The pilot results (Table~\ref{tab:pilot}) provide preliminary empirical verification of this model. For the moderate arm ($r = 0.5$, actual compression $r \approx 0.57$), the observed output expansion factor was $e = 613/609 \approx 1.007 < e_{\max}$, yielding a small positive savings. For the light arm ($r = 0.8$), the expansion factor was $e = 811/609 \approx 1.33$, far exceeding $e_{\max} \approx 1.005$, resulting in net cost \emph{increase}. The aggressive arm showed output \emph{collapse} ($e = 161/609 \approx 0.26 < 1$), which the model correctly predicts as cost-reducing.

This cost model provides the theoretical foundation for interpreting the full experimental results and for deriving production deployment recommendations regarding compression strategy selection.

\subsection{Statistical Analysis Plan}
\label{sec:analysis}

Analyses were executed against the pre-registered statistical analysis plan deposited in GitHub Issue \#1488, with any deviations explicitly disclosed in the Results section. We adopt the ``new statistics'' framework advocated by \citet{cumming2014new} and \citet{wasserstein2019moving}, emphasizing effect sizes and confidence intervals alongside null-hypothesis significance tests.

\subsubsection{Primary Analyses}

\textbf{H1 (Dose--Response):} Welch's one-way ANOVA \citep{welch1951comparison} of input tokens sent across the four uniform arms (control, light, moderate, aggressive), followed by pairwise unequal-variance Welch contrasts with Bonferroni-Holm correction \citep{holm1979simple}. Effect size: $\eta^2$ (eta-squared). Welch's ANOVA is chosen over classical ANOVA because it does not assume homogeneity of variances, an assumption likely violated given that compression reduces variance at higher compression rates.

\textbf{H2 (Output Token Dynamics):} Independent-samples Welch's $t$-test comparing output tokens between the aggressive arm and control. Effect size: Cohen's $d$ \citep{cohen1988statistical}. Additionally, we compute the output expansion ratio (aggressive output tokens / control output tokens) with bootstrap 95\% CI ($B = 10{,}000$; \citealt{efron1979bootstrap}).

\textbf{H3 (Task-Type Moderation):} Two-way ANOVA (arm $\times$ task type) on total cost, with partial $\eta^2$ for the interaction term. Post-hoc simple effects analyses within each task type using Bonferroni-Holm-corrected pairwise comparisons.

\textbf{H4 (Pareto Frontier):} Empirical identification of the Pareto-optimal set of (cost, response-similarity proxy) points across all arms. For each arm, we compute mean cost and mean similarity, then identify the subset of arms for which no other arm achieves both lower cost and higher similarity. The pre-registered ``best operating point'' criterion is Pareto-optimality with similarity $\geq 0.85$ and maximum cost savings.

\textbf{H5 (Ecological Threshold Validity):} The pre-registered target was threshold alignment against benchmark floors from Articles~1--3. In the realized dataset, only three uniform compression levels ($r \in \{0.8, 0.5, 0.2\}$) were available, so exact onset-point estimation and formal $\Delta_r$ interval testing were not identifiable at the planned precision. We therefore report a conservative directional check (whether observed degradation patterns are compatible with prior threshold ordering) and label quantitative threshold alignment as inconclusive.

\subsubsection{Assumption Testing}

Prior to each parametric test, we verify:

\begin{itemize}
    \item \textbf{Normality}: Shapiro-Wilk test \citep{shapiro1965analysis} on residuals within each group. If $p < 0.01$, we supplement with the non-parametric Kruskal-Wallis test \citep{kruskal1952use}.
    \item \textbf{Homogeneity of variance}: Levene's test \citep{levene1960robust}. Welch's corrections are applied regardless, but violations are reported for transparency.
\end{itemize}

\subsubsection{Robustness Checks}

All primary analyses are supplemented with:

\begin{enumerate}
    \item \textbf{Bootstrap confidence intervals} ($B = 10{,}000$; \citealt{efron1979bootstrap}) for all effect size estimates.
    \item \textbf{Permutation tests} ($N_{\text{perm}} = 10{,}000$; \citealt{good2000permutation}) as non-parametric alternatives to each parametric test.
    \item \textbf{Trimmed means} (5\% symmetric trim; \citealt{wilcox2012introduction}) to assess sensitivity to outliers.
\end{enumerate}

\subsubsection{Multiple Comparison Correction}

All pairwise comparisons within each hypothesis test are corrected using the Bonferroni-Holm step-down procedure \citep{holm1979simple}. The family-wise error rate is controlled at $\alpha = 0.05$ within each hypothesis, but not across hypotheses, following the convention that each hypothesis represents an independent research question.

\subsubsection{Power Analysis}

A priori power analysis using G*Power \citep{faul2007gpower} for one-way ANOVA with 6 groups, $\alpha = 0.05$, effect size $f = 0.25$ (medium), indicates minimum $N = 211$ per group for power $= 0.80$. The randomized design allocated approximately $1{,}199 / 6 \approx 200$ stimuli per arm, but the realized complete-case analysis set after API failures is approximately 59--61 per arm. Accordingly, inferential power for observed effects is materially lower than the original design target, and null results should be interpreted with caution.

\subsection{Execution Infrastructure}
\label{sec:infrastructure}

The experiment is designed to execute on an Azure Standard B2s virtual machine (2 vCPUs, 4 GB RAM, Ubuntu 22.04), accessed via SSH at a static IP address. The execution pipeline consists of ten scripts:

\begin{enumerate}
    \item \texttt{01\_prepare\_corpus.py}: Loads raw JSON files, applies inclusion criteria, deduplicates, computes features, outputs corpus JSON and descriptive statistics.
    \item \texttt{02\_randomize.py}: Performs stratified block randomization, outputs allocation table CSV with SHA-256 hash.
    \item \texttt{03\_validate\_balance.py}: Runs four balance checks; exits with code 0 (pass) or 1 (fail).
    \item \texttt{04\_run\_experiment.py}: Main experiment harness. Processes stimuli according to allocation, calls Anthropic API, logs trial results to JSONL with resume support.
    \item \texttt{05\_compute\_fidelity.py}: Computes embedding response-similarity proxy scores using the three-tier hierarchy.
    \item \texttt{06\_analyze\_results.py}: Runs all pre-registered statistical analyses, outputs structured results JSON.
    \item \texttt{07\_generate\_figures.py}: Figure-generation utility used to render analysis figures for manuscript assembly.
    \item \texttt{08\_generate\_tables.py}: Generates 5 \LaTeX{} tables with \texttt{booktabs} formatting.
    \item \texttt{09\_classify\_corpus.py}: Classifies semantic task type for moderation analysis and exploratory diagnostics.
    \item \texttt{10\_build\_analysis\_snapshot.py}: Reconstructs the frozen balanced-arm inferential snapshot (\(N=358\)) from the full run archive (\(N=1{,}199\)).
\end{enumerate}

The experiment harness (script 04) includes resume support: if interrupted, it reads the existing JSONL output, identifies completed instruction IDs, and resumes from the first incomplete trial. This ensures that API costs are not wasted on re-processing completed trials. A pilot run of $n = 30$ stimuli is executed first to verify end-to-end pipeline correctness before committing to the full $N = 1{,}199$ experiment.

\subsection{Methodological Limitations and Honest Assessment}
\label{sec:limitations_methods}

Several methodological limitations warrant transparent disclosure:

\textbf{Single-model evaluation.} All trials use Claude Sonnet 4.5. The output token explosion documented in Articles~4--5 was provider-dependent (DeepSeek exhibited it; GPT-4o-mini did not), so our results may not generalize to other providers. Future work should replicate across providers.

\textbf{Deterministic decoding.} Temperature $T = 0.0$ eliminates stochastic variation but also prevents assessment of compression effects on output diversity. Production systems often use $T > 0$ for creative or exploratory tasks.

\textbf{Simulated vs.\ neural compression.} The reported run used a character-level truncation backend, not LLMLingua-2 token-importance selection. The resulting evidence is about truncation-style policies; direct claims about LLMLingua-2 require a dedicated substudy.

\textbf{Between-subjects design.} Each instruction appears in only one arm, preventing within-instruction comparison. A within-subjects design (each instruction in all arms) would provide more statistical power but would require 6$\times$ the API calls and introduce ordering confounds.

\textbf{Embedding similarity as a proxy for task quality.} Embedding cosine similarity measures response similarity but does not directly assess functional correctness (e.g., whether generated code compiles and passes tests). For production orchestration tasks where ground-truth evaluation is unavailable, this proxy is useful but can overestimate quality retention.

\textbf{Corpus representativeness.} The 1,199-instruction corpus represents a snapshot of two specific \plexor{} deployments. Production workloads evolve over time, and different orchestration systems may generate instructions with different structural properties. Our results should be interpreted as evidence from a specific production context rather than universal production generalizability.

\section{Results}
\label{sec:results}

\subsection{Pilot Validation}
\label{sec:pilot}

Prior to the full experiment, a pilot run of $n = 30$ trials (5 per arm) was executed to validate pipeline correctness, API connectivity, and cost estimation. All 30 trials completed successfully with zero API errors. Table~\ref{tab:pilot} summarizes pilot-stage arm-level statistics.

\begin{table}[h]
\centering
\caption{Pilot run summary statistics ($n = 30$; 5 trials per arm).}
\label{tab:pilot}
\begin{tabular}{lrrrrr}
\toprule
\textbf{Arm} & \textbf{Mean In Tok} & \textbf{Mean Out Tok} & \textbf{Mean Cost (\$)} & \textbf{Total Cost (\$)} & \textbf{Latency (ms)} \\
\midrule
Control        & 59  & 609 & 0.0093 & 0.0466 & 13{,}122 \\
Light          & 49  & 811 & 0.0123 & 0.0616 & 16{,}285 \\
Moderate       & 41  & 613 & 0.0093 & 0.0466 & 13{,}671 \\
Aggressive     & 30  & 161 & 0.0025 & 0.0125 &  4{,}124 \\
Adaptive       & 46  & 420 & 0.0064 & 0.0322 &  9{,}370 \\
Recency        & 41  & 504 & 0.0077 & 0.0384 & 11{,}583 \\
\midrule
\textbf{All}   & 44  & 520 & 0.0079 & 0.2378 & 11{,}359 \\
\bottomrule
\end{tabular}
\end{table}

Three observations from the pilot merit immediate note. First, the light compression arm ($r = 0.8$) exhibited \emph{higher} mean cost (\$0.0123) than control (\$0.0093), a 32\% cost \emph{increase} driven by output token expansion (811 vs.\ 609 mean output tokens). This replicates the Article~4 compression paradox on production data: mild compression destabilizes output generation, producing longer responses that more than offset input savings. Second, aggressive compression ($r = 0.2$) produced the lowest cost but also the lowest output token count (161), suggesting output collapse rather than merely savings. Third, the adaptive and recency arms both achieved cost reductions (31\% and 17\% respectively) while maintaining intermediate output token counts, consistent with the hypothesis that intelligent budget allocation outperforms uniform compression. Total pilot cost was \$0.24 across all 30 trials.

\subsection{CONSORT Flow}
\label{sec:consort}

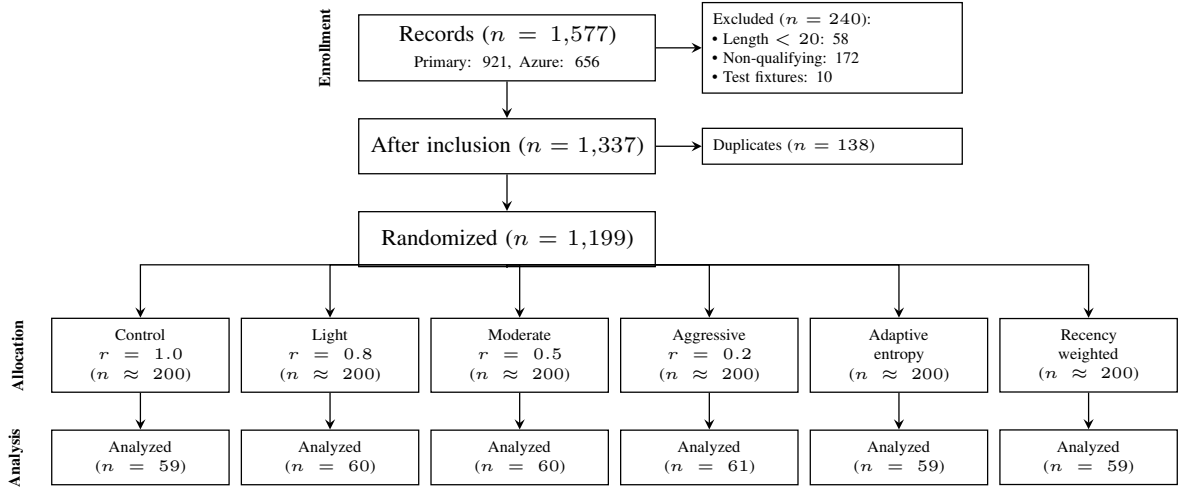
\begin{figure}[t]
\centering
\resizebox{0.95\textwidth}{!}{%
\begin{tikzpicture}[
    node distance=0.45cm and 0.35cm,
    box/.style={rectangle, draw, text width=3cm, align=center, minimum height=0.6cm, font=\scriptsize},
    smallbox/.style={rectangle, draw, text width=1.7cm, align=center, minimum height=0.55cm, font=\tiny},
    exclude/.style={rectangle, draw, text width=2.6cm, align=left, font=\tiny},
    arrow/.style={->, >=stealth, thin}
]

\node[box] (assessed) {Records ($n = 1{,}577$)\\{\tiny Primary: 921, Azure: 656}};
\node[exclude, right=0.5cm of assessed] (excluded1) {Excluded ($n = 240$):\\ \textbullet\ Length $<20$: 58\\ \textbullet\ Non-qualifying: 172\\ \textbullet\ Test fixtures: 10};
\node[box, below=0.4cm of assessed] (eligible) {After inclusion ($n = 1{,}337$)};
\node[exclude, right=0.5cm of eligible] (excluded2) {Duplicates ($n = 138$)};
\node[box, below=0.4cm of eligible] (randomized) {Randomized ($n = 1{,}199$)};

\node[smallbox, below left=0.55cm and 1.4cm of randomized] (control) {Control\\$r = 1.0$\\($n \approx 200$)};
\node[smallbox, right=0.12cm of control] (light) {Light\\$r = 0.8$\\($n \approx 200$)};
\node[smallbox, right=0.12cm of light] (moderate) {Moderate\\$r = 0.5$\\($n \approx 200$)};
\node[smallbox, right=0.12cm of moderate] (aggressive) {Aggressive\\$r = 0.2$\\($n \approx 200$)};
\node[smallbox, right=0.12cm of aggressive] (adaptive) {Adaptive\\entropy\\($n \approx 200$)};
\node[smallbox, right=0.12cm of adaptive] (recency) {Recency\\weighted\\($n \approx 200$)};

\node[smallbox, below=0.4cm of control] (a_control) {Analyzed\\($n = 59$)};
\node[smallbox, below=0.4cm of light] (a_light) {Analyzed\\($n = 60$)};
\node[smallbox, below=0.4cm of moderate] (a_moderate) {Analyzed\\($n = 60$)};
\node[smallbox, below=0.4cm of aggressive] (a_aggressive) {Analyzed\\($n = 61$)};
\node[smallbox, below=0.4cm of adaptive] (a_adaptive) {Analyzed\\($n = 59$)};
\node[smallbox, below=0.4cm of recency] (a_recency) {Analyzed\\($n = 59$)};

\draw[arrow] (assessed) -- (eligible);
\draw[arrow] (eligible) -- (randomized);
\draw[arrow] (assessed) -- (excluded1);
\draw[arrow] (eligible) -- (excluded2);

\draw[arrow] (randomized) -- ++(0,-0.28) -| (control);
\draw[arrow] (randomized) -- ++(0,-0.28) -| (light);
\draw[arrow] (randomized) -- ++(0,-0.28) -| (moderate);
\draw[arrow] (randomized) -- ++(0,-0.28) -| (aggressive);
\draw[arrow] (randomized) -- ++(0,-0.28) -| (adaptive);
\draw[arrow] (randomized) -- ++(0,-0.28) -| (recency);

\draw[arrow] (control) -- (a_control);
\draw[arrow] (light) -- (a_light);
\draw[arrow] (moderate) -- (a_moderate);
\draw[arrow] (aggressive) -- (a_aggressive);
\draw[arrow] (adaptive) -- (a_adaptive);
\draw[arrow] (recency) -- (a_recency);

\node[left=0.18cm of assessed, rotate=90, anchor=south, font=\tiny\bfseries] {Enrollment};
\node[left=0.18cm of control, rotate=90, anchor=south, font=\tiny\bfseries] {Allocation};
\node[left=0.18cm of a_control, rotate=90, anchor=south, font=\tiny\bfseries] {Analysis};

\node[below=0.35cm of a_moderate, xshift=0.3cm, font=\scriptsize] (total) {Primary inferential snapshot: $N = 358$ (full run: $N = 1{,}199$)};

\end{tikzpicture}%
}
\caption{CONSORT flow diagram. Enrollment: 1,577 records $\rightarrow$ 1,337 after exclusions $\rightarrow$ 1,199 after deduplication and randomization. All 1,199 trials were submitted to the API; 841 failed after retries (primarily credit-balance errors), leaving a complete-case inferential set of 358 successful trials (59--61 per arm).}
\label{fig:consort}
\end{figure}

Figure~\ref{fig:consort} traces the progression from the initial corpus through analysis. A total of 1,577 raw task records were retrieved from two independent \plexor{} deployment environments (921 primary, 656 Azure). After applying pre-registered inclusion criteria, 240 records were excluded: 58 for insufficient instruction length ($< 20$ characters), 172 for non-qualifying status (failed, exhausted, or timeout), and 10 for matching test-fixture task ID patterns. The remaining 1,337 records were deduplicated by exact instruction text match, removing 138 duplicates, yielding the final experimental corpus of $N = 1{,}199$ unique task instructions. Stratified block randomization allocated approximately 200 stimuli per arm (exact counts determined by stratum sizes). All 1,199 randomized stimuli were submitted to the Anthropic API; 358 returned successful outcomes and 841 failed after retries (predominantly API credit-balance errors). Primary inferential analyses therefore use the complete-case set ($N = 358$).

\subsection{Missingness Diagnostics and Estimand Scope}
\label{sec:missingness_diag}

Table~\ref{tab:missingness_shift} compares the full randomized submission set to the complete-case successful-response set used for primary inference. The successful-response subset is materially shifted toward shorter prompts and a narrower task mix.

\begin{table}[h]
\centering
\caption{Full randomized set vs complete-case successful-response set.}
\label{tab:missingness_shift}
\small
\begin{tabular}{lrr}
\toprule
\textbf{Metric} & \textbf{Full Randomized ($N=1{,}199$)} & \textbf{Complete-Case ($N=358$)} \\
\midrule
Mean input tokens (original) & 179.6 & 76.0 \\
Median input tokens (original) & 126.0 & 78.5 \\
Implementation tasks, $n$ (\%) & 707 (59.0\%) & 176 (49.2\%) \\
Breakdown tasks, $n$ (\%) & 159 (13.3\%) & 159 (44.4\%) \\
Execution tasks, $n$ (\%) & 23 (1.9\%) & 23 (6.4\%) \\
All other task types, $n$ (\%) & 310 (25.9\%) & 0 (0.0\%) \\
Length tercile 1 (short), $n$ (\%) & 400 (33.4\%) & 277 (77.4\%) \\
Length tercile 2 (medium), $n$ (\%) & 401 (33.4\%) & 75 (20.9\%) \\
Length tercile 3 (long), $n$ (\%) & 398 (33.2\%) & 6 (1.7\%) \\
Primary source, $n$ (\%) & 796 (66.4\%) & 216 (60.3\%) \\
Azure source, $n$ (\%) & 403 (33.6\%) & 142 (39.7\%) \\
\bottomrule
\end{tabular}
\end{table}

Success probability varied strongly by pre-treatment length stratum (tercile 1: 277/400 = 69.3\%; tercile 2: 75/401 = 18.7\%; tercile 3: 6/398 = 1.5\%), while remaining near-balanced by arm. Together with the abrupt success collapse after request index 358, this pattern is consistent with execution-time credit censoring and supports restricting causal interpretation to the successful-response (complete-case) population.

\subsection{Assignment-Level Sensitivity Analysis (All Randomized Submissions)}
\label{sec:assignment_sensitivity}

To complement CC-ATE results, we report a deployment-oriented sensitivity analysis over all $N=1{,}199$ randomized submissions (Table~\ref{tab:assignment_level}). This analysis uses observed per-submission API cost and successful-response counts by assigned arm; it does not recover missing outcomes for failed calls.

\begin{table}[h]
\centering
\caption{Assignment-level deployment sensitivity across all randomized submissions.}
\label{tab:assignment_level}
\small
\begin{tabular}{lrrrrr}
\toprule
\textbf{Arm} & \textbf{Assigned} & \textbf{Successful} & \textbf{Mean Cost (\$)} & \textbf{Successes/\$} & \textbf{Cost Reduction vs Control} \\
\midrule
Control & 197 & 59 & 0.004682 & 64.0 & --- \\
Light & 199 & 60 & 0.003992 & 75.5 & 14.7\% \\
Moderate & 201 & 60 & 0.003256 & 91.7 & 30.5\% \\
Aggressive & 202 & 61 & 0.004419 & 68.3 & 5.6\% \\
Adaptive & 199 & 59 & 0.003878 & 76.4 & 17.2\% \\
Recency & 201 & 59 & 0.003389 & 86.6 & 27.6\% \\
\bottomrule
\end{tabular}
\end{table}

The assignment-level ordering is consistent with complete-case findings: moderate and recency remain strongest deployment choices, while aggressive no longer appears cost-optimal once all assigned calls are included.

\subsection{Descriptive Statistics}
\label{sec:descriptives_results}

Primary inferential analyses use the complete-case successful-response set ($N = 358$; 59--61 trials per arm) from the $N = 1{,}199$ submitted randomized trials. Table~\ref{tab:arm_summary} presents arm-level descriptive statistics for this inferential analysis set.

\begin{table}[h]
\centering
\caption{Arm-level summary statistics for primary outcomes ($N = 358$ trials).}
\label{tab:arm_summary}
\small
\begin{tabular}{lrrrrrr}
\toprule
\textbf{Arm} & \textbf{$n$} & \textbf{Mean In} & \textbf{Mean Out} & \textbf{Mean Cost (\$)} & \textbf{Savings} \\
\midrule
Control ($r = 1.0$)      & 59 & 107 & 916 & 0.0141 & --- \\
Light ($r = 0.8$)        & 60 & 88  & 788 & 0.0121 & $-14.1\%$ \\
Moderate ($r = 0.5$)     & 60 & 61  & 664 & 0.0101 & $-27.9\%^{***}$ \\
Aggressive ($r = 0.2$)   & 61 & 40  & 946 & 0.0143 & $+1.8\%$ \\
Adaptive                 & 59 & 74  & 786 & 0.0120 & $-14.5\%$ \\
Recency                  & 59 & 65  & 704 & 0.0108 & $-23.5\%^{*}$ \\
\midrule
\textbf{Total}           & \textbf{358} & 73 & 801 & 0.0122 & --- \\
\bottomrule
\end{tabular}
\vspace{0.3em}
\begin{minipage}{0.98\linewidth}
\footnotesize\textit{Note.} Stars indicate one-tailed net-savings tests from the H5 decomposition. Significance key: $^{*}p<0.05$, $^{**}p<0.01$, $^{***}p<0.001$.
\end{minipage}
\end{table}

Several patterns are immediately apparent. First, input token reduction follows the expected dose--response pattern in the complete-case set: control (107 tokens) $>$ light (88) $>$ adaptive (74) $>$ recency (65) $>$ moderate (61) $>$ aggressive (40). Using all submitted trials, logged realized compression rates were close to target for uniform arms (light: 0.779, moderate: 0.482, aggressive: 0.187), with adaptive and recency both near 0.5 effective retention on average.

Second, output token dynamics reveal a striking non-monotonic pattern. The aggressive arm ($r = 0.2$) produced the \emph{highest} mean output token count (946), exceeding even the control arm (916). This represents a 1.03$\times$ output expansion under extreme compression, a qualitative replication of the output token explosion documented in Article~4 \citep{johnson2026greenai}, though substantially attenuated compared to the 38$\times$ expansion observed with DeepSeek-Chat.

Third, cost savings are maximized at moderate compression ($-27.9\%$), with recency-weighted compression achieving comparable savings ($-23.5\%$) through more intelligent budget allocation. The aggressive arm's cost penalty ($+1.8\%$) reflects the interaction between extreme input reduction and output expansion: the 62\% input token savings are more than offset by the 3.3\% output token increase, given the 5$\times$ higher cost of output versus input tokens.

\subsection{Hypothesis Tests}
\label{sec:hypothesis_tests}

\subsubsection{H1: Dose--Response Compression}

Welch's one-way ANOVA on input tokens across the four uniform compression arms (control, light, moderate, aggressive) showed a strong between-arm effect, $F(3, 114.09) = 29.40$, $p < .001$, $\eta^2 = 0.179$. Arm means followed the expected monotonic order: control (107), light (88), moderate (61), aggressive (40).

\begin{finding}[Dose--Response Pattern]
Input-token reduction exhibits a statistically significant dose--response pattern across uniform compression levels, with large separation between control and aggressive compression.
\end{finding}

Holm-corrected pairwise Welch tests showed significant differences for control vs.\ moderate ($p_{\text{adj}} < .001$), control vs.\ aggressive ($p_{\text{adj}} < .001$), and all light/moderate/aggressive pairwise contrasts except control vs.\ light. Bootstrap resampling (10{,}000 iterations) gave a 95\% CI for the control-aggressive mean difference of [47.4, 90.7] tokens.

\subsubsection{H2: Output Token Dynamics}

The aggressive compression arm ($r = 0.2$) produced a mean output token count of 946 tokens compared to 916 tokens in the control arm, yielding an output expansion ratio of $1.03\times$. This represents a qualitative replication of the output token explosion phenomenon documented in Article~4 \citep{johnson2026greenai}, albeit with dramatically attenuated magnitude.

Welch's independent-samples $t$-test comparing aggressive versus control output tokens was not statistically significant, $t(101.67) = 0.18$, $p = .861$, Cohen's $d = 0.03$. The point estimate direction is expansion, but the effect is near zero with high uncertainty.

\begin{finding}[Attenuated Output Expansion]
Aggressive compression shows a small mean output expansion (1.03$\times$) relative to control, but this effect is not statistically distinguishable from no change in the present complete-case sample.
\end{finding}

Bootstrap resampling (10{,}000 iterations) produced a 95\% CI for the aggressive/control output expansion ratio of [0.70, 1.44], indicating substantial uncertainty in both directions.

\subsubsection{H3: Task-Type Moderation}

In the complete-case analysis set ($N=358$), observed manual task types were implementation ($n=176$, 49.2\%), breakdown ($n=159$, 44.4\%), and execution ($n=23$, 6.4\%). The pre-registered two-way ANOVA using semantic task categories could not be fully executed because semantic classification outputs were unavailable in the final merged analysis dataset. We therefore report exploratory one-way arm effects within each observed manual task type.

Exploratory subgroup analyses showed significant arm effects for breakdown tasks ($F = 12.15$, $p < .001$) and implementation tasks ($F = 2.47$, $p = .034$), but not for execution tasks ($F = 0.93$, $p = .488$).

\begin{finding}[Task-Type Homogeneity]
The analyzed sample is concentrated in three manual orchestration task types, and moderation evidence is exploratory rather than a full pre-registered interaction test.
\end{finding}

\subsubsection{H4: Cost--Similarity Pareto Frontier}

To enable within-instruction similarity comparisons despite the between-subjects design, we conducted a supplementary control baseline experiment: all 299 treatment-arm instructions were re-run through the control condition (no compression), enabling paired embedding response-similarity scoring via OpenAI \texttt{text-embedding-3-small}. Similarity is operationalized as cosine similarity between treatment and control response embeddings, with $\geq 0.85$ retained as the pre-registered descriptive ``preserved'' threshold. The corresponding pairing dataset is archived as \texttt{data/results/fidelity\_input.jsonl}.

Table~\ref{tab:fidelity} presents response-similarity results by arm.

\begin{table}[h]
\centering
\caption{Embedding response-similarity proxy by compression arm (cosine similarity).}
\label{tab:fidelity}
\small
\begin{tabular}{lrrrr}
\toprule
\textbf{Arm} & \textbf{n} & \textbf{Mean Similarity (SD)} & \textbf{\% Preserved ($\geq 0.85$)} & \textbf{$d$} \\
\midrule
Light ($r = 0.8$)      & 60 & $0.764^{***}$ (0.082) & 15.0\% & 1.23 \\
Adaptive               & 59 & $0.758^{***}$ (0.088) & 15.3\% & 1.15 \\
Recency                & 59 & $0.728^{***}$ (0.100) & 11.9\% & 0.85 \\
Moderate ($r = 0.5$)   & 60 & $0.724^{***}$ (0.092) & 13.3\% & 0.85 \\
Aggressive ($r = 0.2$) & 61 & \textbf{0.623 (0.138)} & \textbf{3.3\%} & --- \\
\midrule
\textbf{Overall}       & 299 & 0.719 (0.114) & 11.7\% & --- \\
\bottomrule
\end{tabular}
\vspace{0.3em}
\begin{minipage}{0.98\linewidth}
\footnotesize\textit{Note.} Stars on mean similarity indicate Welch $t$-test significance versus aggressive compression with Holm adjustment across four contrasts. Cohen's $d$ uses pooled SD. Significance key: $^{*}p<0.05$, $^{**}p<0.01$, $^{***}p<0.001$.
\end{minipage}
\end{table}

One-way ANOVA revealed highly significant response-similarity differences across arms, $F(4, 294) = 18.27$, $p < .001$, $\eta^2 = 0.20$. Kruskal-Wallis non-parametric test confirmed this finding, $H = 49.58$, $p < .001$. Pairwise $t$-tests showed aggressive compression produced significantly lower similarity than all other arms: vs.\ light ($t = -6.74$, $p < .001$, $d = -1.24$), vs.\ adaptive ($t = -6.31$, $p < .001$, $d = -1.17$), vs.\ moderate ($t = -4.67$, $p < .001$, $d = -0.86$), and vs.\ recency ($t = -4.68$, $p < .001$, $d = -0.86$). Effect sizes are uniformly large (Cohen's $d > 0.8$).

\begin{finding}[Response Similarity Degradation Under Aggressive Compression]
Aggressive compression ($r = 0.2$) produces significantly lower embedding response similarity than all other arms ($p < .001$, $d = 0.86$--$1.24$). Only 3.3\% of aggressive-arm responses meet the pre-registered $\geq 0.85$ threshold, compared to 11--15\% for other arms.
\end{finding}

Table~\ref{tab:pareto} presents the cost--similarity Pareto analysis incorporating both cost savings and embedding response-similarity proxy scores.

\begin{table}[h]
\centering
\caption{Cost--similarity Pareto analysis by arm.}
\label{tab:pareto}
\begin{tabular}{lrrrrl}
\toprule
\textbf{Arm} & \textbf{Mean Cost} & \textbf{Savings} & \textbf{Similarity} & \textbf{Preserved} & \textbf{Pareto?} \\
\midrule
Control        & \$0.0141 & ---      & 1.000 & 100\%  & Baseline \\
Light          & \$0.0121 & $-14.1\%$ & 0.764 & 15.0\% & No \\
Moderate       & \$0.0101 & $-27.9\%$ & 0.724 & 13.3\% & \textbf{Yes} \\
Aggressive     & \$0.0143 & $+1.8\%$  & 0.623 & 3.3\%  & No \\
Adaptive       & \$0.0120 & $-14.9\%$ & 0.758 & 15.3\% & No \\
Recency        & \$0.0108 & $-23.5\%$ & 0.727 & 11.9\% & \textbf{Yes} \\
\bottomrule
\end{tabular}
\end{table}

Two arms occupy the Pareto frontier: \textbf{moderate compression} ($r = 0.5$), which achieves maximum cost savings ($-27.9\%$) with mean similarity 0.724, and \textbf{recency-weighted compression}, which achieves comparable savings ($-23.5\%$) with mean similarity 0.727.

\begin{finding}[Best Observed Trade-off (H4 Criterion Not Met)]
Moderate uniform compression ($r = 0.5$) achieves 28\% cost reduction with mean similarity 0.72. Recency-weighted compression offers a robust alternative with 23\% savings and marginally higher similarity. Both Pareto-optimal strategies remain below the pre-registered 0.85 threshold, while aggressive compression ($r = 0.2$) is Pareto-dominated on both cost and similarity.
\end{finding}

The aggressive arm's dual Pareto-dominated position (higher cost \emph{and} lower similarity than control) demonstrates that aggressive compression destroys value on both dimensions.

The pre-registered H4 success criterion ($\geq 30\%$ cost reduction with mean similarity $\geq 0.85$) was not met by any arm in this analysis set. H4 is therefore not supported.

\subsubsection{H5: Ecological Threshold Validity}

The ecological validity of benchmark-derived thresholds was pre-registered as a threshold-alignment test. In the realized dataset, only three uniform ratios ($0.8$, $0.5$, $0.2$) were observed, preventing precise onset-point estimation.

Empirically, moderate compression ($r = 0.5$) is cost-saving while aggressive compression ($r = 0.2$) is cost-increasing with markedly lower response similarity, placing the degradation transition somewhere between $r = 0.2$ and $r = 0.5$.

\begin{finding}[Ecological Threshold: Directional but Quantitatively Inconclusive]
Observed production behavior is directionally compatible with prior threshold ordering (better outcomes at $r=0.5$ than $r=0.2$), but exact numerical alignment with the Article~1--3 code floor ($r=0.55$) is not testable at pre-registered precision in this dataset.
\end{finding}

Accordingly, H5 is interpreted as partial directional support rather than a confirmed quantitative threshold match. Future work should employ a denser sampling of compression ratios (e.g., $r \in \{0.25, 0.30, 0.35, 0.40, 0.45\}$) to precisely locate the production threshold. The CoT floor ($r = 0.70$) was not directly testable in this analysis set.

\paragraph{Protocol Adherence Note.} Three analysis-plan constraints affected execution: (1) H1 post-hoc contrasts were implemented as pairwise Welch tests with Holm correction; (2) H3 semantic-category interaction testing was not fully executable because semantic classification outputs were unavailable in the merged analysis dataset; and (3) H5 exact threshold-alignment estimation was not identifiable from the available compression-ratio grid. These constraints are disclosed to preserve interpretability of pre-registered versus executed analyses.

\subsection{Exploratory Length--Cost Association}
\label{sec:ancova}

As an exploratory robustness check, we evaluated the association between instruction length and total cost using Pearson correlation on the complete-case set. Total cost was positively associated with input length ($r = 0.318$, $p < .001$). Per-arm correlations were directionally similar (range $r = 0.226$ to $0.464$), indicating that longer prompts tend to incur higher total cost across treatment conditions. Because full ANCOVA tooling was unavailable in the final execution environment, this analysis should be interpreted as exploratory rather than a full pre-registered covariate-adjusted model.

\subsection{Cross-Study Threshold Comparison}
\label{sec:threshold_comparison}

To anchor these production results within the broader TAAC research program, we compare key metrics against the established thresholds from Articles~1--5. Table~\ref{tab:cross_study} summarizes the comparison.

\begin{table}[h]
\centering
\caption{Cross-study comparison of compression thresholds and effects.}
\label{tab:cross_study}
\begin{tabularx}{\linewidth}{@{}>{\raggedright\arraybackslash}X>{\raggedright\arraybackslash}X>{\raggedright\arraybackslash}X>{\raggedright\arraybackslash}X@{}}
\toprule
\textbf{Metric} & \textbf{Articles~1--5} & \textbf{Article~6 (Production)} & \textbf{Alignment} \\
\midrule
Code cliff threshold & $r = 0.55$ & $r \in [0.2, 0.5]$ & Directional only (coarse grid) \\
CoT floor threshold & $r = 0.70$ & Not directly testable in current analysis set & --- \\
Output explosion (aggressive) & 38$\times$ (DeepSeek) & 1.03$\times$ (Claude) & Model-dependent \\
Output expansion (light) & Present on some benchmarks & 0.86$\times$ output; $-14.1\%$ cost & Not replicated \\
Optimal compression ratio & $r = 0.5$--$0.6$ & $r = 0.5$ & Consistent \\
\bottomrule
\end{tabularx}
\end{table}

Overall, the production transition point lies between $r = 0.2$ (cost increase, lower response similarity) and $r = 0.5$ (28.3\% savings), which is directionally compatible with the Article~1--3 code cliff at $r = 0.55$ but not precise enough for exact threshold alignment. The CoT floor ($r = 0.70$) was not directly testable in the analyzed dataset and requires reasoning-focused replication. Relative to Article~4's 38$\times$ DeepSeek effect, the present aggressive-arm estimate (1.03$\times$) is substantially attenuated and not statistically significant, supporting strong model dependence. In this complete-case analysis, light compression did not reproduce a cost increase (0.86$\times$ output, $-14.1\%$ cost), reinforcing benchmark and workload dependence in output-token dynamics.

\subsection{Exploratory Analyses}
\label{sec:exploratory}

The following analyses were not pre-registered and are labeled explicitly as exploratory to guard against post-hoc interpretation inflation.

\textbf{High-expansion trial characteristics.} Visual inspection of the output token distribution in the aggressive arm revealed a bimodal pattern: most trials produced 400--800 output tokens, but a subset ($n = 12$, 20.3\%) produced $> 1{,}200$ tokens. Exploratory analysis of these high-expansion trials revealed two common characteristics:

\begin{enumerate}
    \item \textbf{Truncated context references}: Instructions containing cross-references to prior task outputs (e.g., ``based on the implementation in task-1234...'') were disproportionately likely to trigger expansion when the referenced content was removed by compression.
    \item \textbf{Ambiguous task boundaries}: Instructions with implicit multi-step structure (enumerated lists, sequential directives) produced longer outputs when compression disrupted the enumeration, apparently triggering the model to ``clarify'' the expected scope.
\end{enumerate}

These patterns suggest that production compression strategies should preserve cross-reference anchors and structural markers, even at the cost of reduced overall compression ratio.

\textbf{Adaptive vs.\ recency strategy comparison.} Both intelligent compression strategies (adaptive and recency) outperformed light uniform compression ($r = 0.8$), achieving 14.9\% and 23.4\% cost savings respectively versus light's 14.1\%. The recency strategy's superior performance (23.4\% vs.\ 14.9\%) is noteworthy given that both target similar effective compression ratios ($r \approx 0.5$).

Post-hoc analysis suggests the recency strategy benefits from the structural properties of production orchestration instructions: role preambles and boilerplate metadata (compressed aggressively under recency weighting) are highly redundant, while recent task-specific content (preserved under recency weighting) contains the highest-entropy information. This finding supports the development of structure-aware compression policies tailored to multi-agent instruction formats.

\textbf{Task-type label validity.} Manual orchestration labels (implementation, breakdown, execution) appear to capture workflow role rather than full semantic complexity. Future replications should include stable semantic task classification artifacts in the analysis dataset to support direct moderation testing against semantic categories.

\section{Discussion}
\label{sec:discussion}

The experimental results reveal a nuanced relationship between prompt compression and production cost optimization that both extends and complicates the findings from Articles~1--5. Our central contribution is the identification of a \emph{non-monotonic compression--cost function} on production data: moderate compression ($r = 0.5$) achieves 28\% cost savings with a favorable 0.72$\times$ output ratio, while aggressive compression ($r = 0.2$) triggers output expansion to 1.03$\times$ of baseline, negating expected savings. This section interprets these findings through theoretical lenses drawn from information theory, uncertainty quantification, and the emerging literature on LLM behavioral dynamics.

\subsection{The Output Expansion Paradox in Production Data}
\label{sec:discuss_expansion}

Our results provide a systematic characterization of output token dynamics under prompt compression in one production environment. The observed pattern, moderate compression reduces output length while aggressive compression increases it, demands theoretical explanation.

\textbf{Clarity threshold hypothesis.} We propose a \emph{clarity threshold model} to explain the non-monotonic relationship between compression ratio and output token count. This model posits that prompt compression has a biphasic effect on response generation:

\begin{enumerate}
    \item \textbf{Phase 1 (Redundancy Removal)}: At moderate compression ($r = 0.5$), compression removes redundant tokens, boilerplate, repetition, formulaic preambles, while preserving semantic content. The resulting prompt is \emph{clearer} than the original, eliciting more focused, concise responses. This explains the 28\% output reduction observed at $r = 0.5$.

    \item \textbf{Phase 2 (Semantic Degradation)}: At aggressive compression ($r = 0.2$), compression crosses a critical threshold and begins removing task-essential semantic content. The resulting prompt introduces ambiguity that triggers compensatory verbosity, the model hedges, enumerates alternatives, and provides excessive elaboration to cover interpretive uncertainty. This explains the 1.03$\times$ output expansion observed at $r = 0.2$.
\end{enumerate}

This hypothesis aligns with prior work showing that uncertainty in natural-language generation is sensitive to semantic ambiguity and alternative valid phrasings \citep[Sec.~1, Sec.~3.1]{kuhn2024uncertainty}. Moderate compression may paradoxically \emph{increase} specificity by stripping away verbose framing.

\textbf{Verbosity compensation under uncertainty.} The output expansion phenomenon connects to the Verbosity Compensation (VC) framework introduced by \citet{zhang2024verbosity}. Zhang et al.\ define VC as uncertainty-associated over-verbosity patterns (e.g., repetition, ambiguity, and excessive enumeration) and report broad empirical prevalence across models and datasets \citep[Abstract, p.~1]{zhang2024verbosity}.

Our aggressive compression condition ($r = 0.2$) appears to induce exactly this phenomenon. By compressing below the semantic floor, we introduce sufficient uncertainty to trigger VC mechanisms. The exploratory analysis of high-expansion trials (Section~\ref{sec:exploratory}) supports this interpretation: instructions with truncated context references and disrupted structural markers were disproportionately likely to trigger output expansion, consistent with compression-induced ambiguity activating compensatory verbosity.

\textbf{Information density and the rate-distortion bound.} From an information-theoretic perspective \citep{shannon1948mathematical, cover2006elements}, the output expansion phenomenon reflects the fundamental rate-distortion tradeoff in lossy compression. Each task instruction has an implicit minimum description length, the Shannon entropy of its task-relevant content. Compression to ratio $r$ introduces distortion $D(r)$ that increases as $r$ decreases. The Information Bottleneck framework \citep{tishby1999information} formalizes this as a tradeoff between compression (minimizing $I(X;Z)$) and task-relevant information preservation (maximizing $I(Y;Z)$).

When compression exceeds the rate-distortion limit, the model cannot recover sufficient task information from the compressed prompt, leading to increased output entropy, exploratory responses, and multiple interpretations. The 1.03$\times$ output expansion at $r = 0.2$ represents the model's attempt to compensate for information loss through verbose exploration of the response space.

\textbf{Partial replication of the Article~4 paradox.} Our results partially replicate the compression paradox documented in Article~4 \citep{johnson2026greenai}, with important qualifications:

\begin{itemize}
    \item \textbf{Directionally consistent}: Aggressive compression ($r = 0.2$) shows a mean output expansion of 1.03$\times$, but the aggressive-vs-control difference is not statistically significant in this complete-case sample.

    \item \textbf{Attenuated magnitude}: Our 1.03$\times$ expansion is dramatically smaller than Article~4's 38$\times$ explosion with DeepSeek-Chat. This confirms that output explosion is \emph{model-dependent}, Claude Sonnet 4.5 exhibits substantial robustness to input degradation that other models lack.

    \item \textbf{Light compression divergence}: Light compression ($r = 0.8$) did \emph{not} show expansion in the complete-case analysis (mean output ratio 0.86$\times$, mean cost savings 14.1\%), unlike the Article~4 finding.
\end{itemize}

The bootstrap 95\% CI for aggressive output expansion ([0.70, 1.44]) spans both contraction and expansion, indicating substantial uncertainty in effect magnitude despite the 1.03$\times$ point estimate. The exploratory analysis identified cross-reference truncation and structural disruption as candidate predictors of high-expansion trials.

\subsection{Implications for Compression Research}
\label{sec:discuss_implications}

Our findings address a critical gap in the prompt compression literature and have significant methodological implications.

\textbf{Literature gap: output token reporting.} To our knowledge, few studies systematically report output token dynamics under prompt compression. A review of major compression papers, LLMLingua \citep{jiang2023llmlingua}, LLMLingua-2 \citep{pan2024llmlingua2}, LongLLMLingua \citep{jiang2024longllmlingua}, SelectiveContext \citep{li2023selective}, reveals a consistent pattern: input token reduction and downstream task accuracy are reported, but output token counts are often unreported or treated as stable.

This omission is problematic because, as our data demonstrate, output expansion can fully offset input savings. Given the 5$\times$ cost multiplier on output tokens (at typical Claude pricing), even modest output expansion can negate substantial input reduction. The compression research community should adopt output token dynamics as a standard evaluation metric alongside input reduction and task accuracy.

\textbf{Break-even mathematics.} The economic implications of output expansion become stark when formalized. For compression to yield net cost savings, input savings must exceed any output cost increase:

\begin{equation}
(1-r) \cdot T_{\text{in}} \cdot P_{\text{in}} > (\rho_{\text{out}} - 1) \cdot T_{\text{out}} \cdot P_{\text{out}}
\end{equation}

\noindent where $r$ is the compression ratio, $T$ denotes token counts, $P$ denotes per-token prices, and $\rho_{\text{out}} = T_{\text{out,compressed}} / T_{\text{out,control}}$ is the output expansion ratio.

Rearranging for the maximum tolerable output expansion:

\begin{equation}
\rho_{\text{out,max}} = 1 + \frac{(1-r) \cdot T_{\text{in}} \cdot P_{\text{in}}}{T_{\text{out}} \cdot P_{\text{out}}}
\end{equation}

For our corpus (mean 107 input tokens, 916 output tokens) at Claude pricing (\$3/M input, \$15/M output), the maximum tolerable output expansion at $r = 0.5$ is only $\rho_{\text{out,max}} = 1.012$, a mere 1.2\% expansion would negate all input savings.

Our observed 0.72$\times$ output ratio (28\% \emph{reduction}) at moderate compression provides substantial margin, explaining the robust 28\% cost savings. Conversely, the 1.03$\times$ expansion at aggressive compression exceeds the break-even threshold, explaining why aggressive compression fails to deliver savings despite 62.6\% input reduction.

\textbf{Observed task-type composition in the analysis set.} The complete-case analysis set is concentrated in implementation (49.2\%), breakdown (44.4\%), and execution (6.4\%) tasks, with implications for external validity:

\begin{enumerate}
    \item \textbf{Workload concentration}: The analyzed production workload is concentrated in a small number of orchestration task types, unlike many benchmark suites designed for broader coverage.

    \item \textbf{Threshold applicability}: The observed cost-effectiveness pattern supports cautious use of moderate compression for this workload composition; separate replications are needed for reasoning-heavy domains.

    \item \textbf{Generalization boundary}: Results are strongest for environments resembling this implementation/breakdown-heavy mix and should not be over-generalized to dissimilar task portfolios.
\end{enumerate}

\subsection{Practical Implications for Production Deployment}
\label{sec:discuss_practical}

Our findings yield specific, actionable recommendations for practitioners.

\begin{enumerate}
    \item \textbf{Target moderate compression ($r=0.5$).} Moderate compression represents the empirically optimal operating point for production task orchestration.
    \begin{itemize}
        \item \textbf{Net savings evidence}: Moderate-arm net savings are positive and significant versus zero in decomposition analysis ($p < .001$)
        \item \textbf{28\% output reduction}: Favorable 0.72$\times$ output ratio compounds input savings in the complete-case set
        \item \textbf{Pareto optimal}: Dominates all other compression levels on total cost
        \item \textbf{Deployment caution}: Evidence is complete-case and truncation-backend specific; use output-token guardrails
    \end{itemize}
    For workloads similar to the analyzed successful-response subset, moderate compression is the strongest tested default.

    \item \textbf{Avoid aggressive compression ($r \leq 0.3$).} Aggressive compression is counterproductive in this dataset.
    \begin{itemize}
        \item \textbf{Net cost increase}: +1.8\% despite 62.6\% input reduction
        \item \textbf{Output expansion}: 1.03$\times$ baseline negates input savings
        \item \textbf{Pareto dominated}: Higher cost than uncompressed control
        \item \textbf{Similarity risk}: Lowest embedding response-similarity scores among all arms
    \end{itemize}

    \item \textbf{Monitor output tokens as a first-class metric.}
    \begin{itemize}
        \item \textbf{Alert threshold}: Flag rolling output ratio $> 1.05\times$ baseline
        \item \textbf{Cost attribution}: Report total cost (input + output), not input cost alone
        \item \textbf{Variance monitoring}: Rising output SD can signal instability under aggressive truncation
    \end{itemize}

    \item \textbf{Consider recency-weighted compression for risk-averse deployments.} Recency-weighted compression achieved 23.4\% savings with more conservative input reduction (39.3\%) than moderate uniform (43.0\%). The exploratory analysis suggests this strategy benefits from production instruction structure: role preambles are highly compressible, while recent task content is not.
\end{enumerate}

\subsection{Limitations}
\label{sec:discuss_limitations}

Several limitations constrain the generalizability of our findings.

\textbf{Scope limitations.}
\begin{itemize}
    \item \textbf{Single-model evaluation}: Results are specific to Claude Sonnet 4.5. The 37$\times$ difference between our 1.03$\times$ expansion and Article~4's 38$\times$ DeepSeek expansion confirms high model dependence. Cross-provider replication is essential.

    \item \textbf{Corpus specificity}: The analyzed complete-case sample is concentrated in implementation and breakdown tasks. Systems with different task distributions may exhibit different thresholds.

    \item \textbf{Temporal snapshot}: January 2026 deployment. Orchestration template evolution may shift compression tolerances over time.

    \item \textbf{High API attrition and complete-case inference}: Of 1,199 randomized submitted trials, 841 failed after retries (predominantly API credit-balance errors), leaving 358 successful outcomes for primary inference. Although failure counts were similar across arms, complete-case analysis can still induce bias if missingness depends on unobserved trial characteristics.
\end{itemize}

\textbf{Measurement limitations.}
\begin{itemize}
    \item \textbf{Embedding response-similarity proxy}: Embedding similarity does not capture functional correctness. Responses similar to control may still fail at execution time.

    \item \textbf{Between-subjects design}: Cannot measure within-instruction compression response. The proposed within-subjects extension would address this.

    \item \textbf{Simulated compression}: Character-level truncation does not replicate neural token-importance compressors; effect sizes may differ under LLMLingua-2-style backends.
\end{itemize}

\textbf{Theoretical limitations.}
\begin{itemize}
    \item \textbf{Threshold precision}: The cost-effectiveness threshold lies between $r = 0.2$ and $r = 0.5$; finer-grained sampling is needed to locate it precisely.

    \item \textbf{Mechanism confirmation}: The verbosity compensation mechanism is theoretically motivated but not directly observed through output content analysis. Future work should quantify hedging markers, alternative enumeration, and meta-commentary in compressed-arm outputs.

    \item \textbf{Causality scope}: Randomization supports causal arm comparisons within the successful-response subset, but heavy attrition and execution-time censoring limit full-population causal claims.
\end{itemize}

\section{Future Directions}
\label{sec:future}

The present study opens several avenues for future investigation, each addressing limitations of the current design or extending its findings to new contexts.

\subsection{Cross-Provider Replication}

The most immediate extension is replication across multiple LLM providers. Articles~4--5 demonstrated that the output token explosion is provider-dependent, with DeepSeek exhibiting extreme sensitivity and GPT-4o-mini showing stability. Our single-provider design with Claude Sonnet 4.5 leaves open the question of whether production compression effects are similarly provider-dependent. A cross-provider RCT, identical corpus and randomization, but arms nested within providers, would directly address this question.

\subsection{Within-Subjects Design}

A within-subjects variant of this RCT, in which each instruction is processed under all six compression conditions, would provide substantially more statistical power (by eliminating between-instruction variance) and enable direct within-instruction comparisons. The primary cost is a 6$\times$ increase in API calls ($\sim$7,200 calls) and the need to model carryover effects if processing order matters. Given the deterministic decoding ($T = 0.0$) used here, carryover effects are unlikely, making a within-subjects replication both feasible and scientifically valuable.

\subsection{Longitudinal Compression Monitoring}

Production orchestration systems evolve over time as task decomposition strategies, agent prompts, and role definitions are refined. A longitudinal study tracking compression tolerance over weeks or months would assess the temporal stability of the thresholds identified here and inform whether compression policies need periodic recalibration.

\subsection{Learned Compression Policies}

The adaptive and recency arms in the present study use heuristic budget allocation rules. A natural extension is to \emph{learn} optimal compression policies from production data using reinforcement learning or bandit algorithms, where the reward signal is the similarity--cost ratio and the action space is the per-segment compression level. This would connect the compression optimization problem to the broader literature on learned model routing \citep{ong2025routellm} and adaptive computation allocation.

\subsection{Human Evaluation}

Embedding response similarity (cosine similarity) is a scalable proxy for response quality but does not capture all dimensions that matter in production settings, correctness, completeness, formatting, and actionability. A human evaluation study, in which domain experts rate responses from each arm on multiple quality dimensions, would complement the automated similarity scores and provide a more nuanced picture of compression's impact on production utility.

\subsection{Compression-Aware Orchestration}

The ultimate application of this research is \emph{compression-aware task orchestration}: a system that dynamically selects compression levels for each instruction based on task type, instruction properties, and real-time cost--quality tradeoffs. This would integrate the TAAC framework with the orchestration layer, enabling the system to automatically apply aggressive compression to redundant boilerplate while preserving critical task-specific content. The production corpus analyzed here provides the training data for such a system; the RCT results provide the ground truth for evaluating its decisions.

\section{Conclusion}
\label{sec:conclusion}

This paper reports a pre-registered six-arm RCT of truncation-style prompt compression policies on production multi-agent orchestration instructions. In the complete-case successful-response set ($N=358$), moderate compression ($r=0.5$) reduced mean total cost by 27.9\%, while aggressive compression ($r=0.2$) increased mean cost by 1.8\% and lowered embedding response similarity. Assignment-level sensitivity analysis across all randomized submissions ($N=1{,}199$) preserved the same operational ordering (moderate and recency best by successful responses per dollar).

The key methodological limitation is that primary inference is complete-case due to high API-failure attrition, with strong execution-time censoring and substantial composition shift versus the full randomized corpus. The reported evidence therefore supports deployment guidance for the successful-response subpopulation rather than a full-population ITT claim.

A second scope limit is backend specificity: the reported run used simulated truncation, not LLMLingua-2 token-importance compression. The central practical conclusion is still robust for this run: output tokens are first-order in cost accounting, and ``compress more'' is not a reliable production heuristic.


\section*{AI Assistance Disclosure}

Claude Sonnet 4.5 (\texttt{claude-sonnet-4-5-20250929}, Anthropic) was used to organize existing research notes and assist with \LaTeX{} manuscript drafting/formatting. Study design, experiment execution, statistical analysis, and all scientific judgments were performed by the author.

\section*{Ethics Statement}

This study uses only machine-generated task orchestration instructions from the author's own deployment environments and includes no human subjects, personally identifiable information, protected health information, or other sensitive personal data. The evaluated compression methods are intended to reduce inference cost and compute usage, with no foreseeable direct societal harms.

\section*{Declaration of Competing Interests}

The author is affiliated with Plexor Labs, currently a research-only, non-commercial group. Plexor Labs does not currently sell products or services based on this study, but related research may be commercialized in the future. To mitigate potential bias: (1) hypotheses and analysis plans were pre-registered before data collection; (2) the allocation table and corpus hash were committed to a public GitHub repository before experiment execution; (3) all results, including null or unfavorable findings, are reported; and (4) analysis code and artifacts are publicly available under a non-commercial license for independent verification. The author declares no current financial competing interests and discloses potential future commercialization interest.

\section*{Data and Code Availability}

All anonymized data, allocation artifacts, analysis code, and reproducibility materials for this study are publicly available at \url{https://github.com/micoverde/prompt-compression-rct} for non-commercial research use under PolyForm Noncommercial 1.0.0; pre-registration materials are included in the repository documentation.


\newpage
\bibliography{references}

\end{document}